\documentclass[12pt, titlepage, reqno]{article}

 \usepackage[cmex10]{amsmath}

 \usepackage{amssymb, latexsym}
 \usepackage{amsmath} 
 \usepackage{amsthm}
 \usepackage{bm}
 \usepackage[mathscr]{eucal}
 \usepackage{enumerate}

 \usepackage{natbib}
 \usepackage[french,english]{babel}

\usepackage{algorithm,algpseudocode}
\usepackage{caption}
\usepackage{color}

\usepackage{gensymb}

\usepackage[caption = false]{subfig}

\usepackage{float}

\usepackage{algpseudocode}

\usepackage{graphicx}

\usepackage{tabularx}
\usepackage{slashbox}

\usepackage{hyperref}
\hypersetup{
    colorlinks,%
    citecolor=blue,%
    filecolor=blue,%
    linkcolor=blue,%
    urlcolor=blue,
}  
  
\usepackage{multirow}

\graphicspath{{./Figures/}}

\begin{document}

 \begin{titlepage}

 \begin{center}

\textbf{Deep scattering transform applied to note onset detection and instrument recognition}

\vspace{10ex}

Cazau Dorian$^{a}$ \footnote{Corresponding author e-mail: cazaudorian@outlook.fr}, Guillaume Revillon$^{a}$ and Olivier Adam$^{a}$\\ 

\vspace{1ex}

\begin{flushleft}

\small{$^a$ Sorbonne Universit\'es, UPMC University Paris 06/CNRS, UMR 7190, Institut Jean le Rond, d'Alembert, F-75015, Paris, France} \\

\end{flushleft}

 \end{center}

 \end{titlepage}

\begin{abstract}
Automatic Music Transcription (AMT) is one of the oldest and most well-studied problems in the field of music information retrieval. Within this challenging research field, onset detection and instrument recognition take important places in transcription systems, as they respectively help to determine exact onset times of notes and to recognize the corresponding instrument sources. The aim of this study is to explore the usefulness of multiscale scattering operators for these two tasks on plucked string instrument and piano music. After resuming the theoretical background and illustrating the key features of this sound representation method, we evaluate its performances comparatively to other classical sound representations. Using both MIDI-driven datasets with real instrument samples and real musical pieces, scattering is proved to outperform other sound representations for these AMT subtasks, putting forward its richer sound representation and invariance properties.
\end{abstract}

\addtocounter{page}{2}

\section{Introduction}

Onset detection and instrument recognition in polyphonic music are two of the most important sub-tasks in Automatic Music Transcription (AMT), and represent processing challenges on their own in the Music Information Retrieval (MIR) community. On one side, onset detection can be roughly defined as the automatic process of locating each single note onset in a musical piece, and where the notion of note onset, perhaps better defined as a phenomenal accent \citep{Lerdahl1983}, refers to discrete temporal events in an audio stream where there is a marked change in any of the perceived psychoacoustical properties of sound, i.e., loudness, timbre, and pitch. Its applications are multifold. Onset detection is a frontend to beat induction algorithms \citep{Klapuri2004d}, empowers segmentation for rhythmic analysis (tempo identification and meter identification) and event manipulation both online and offline \citep{Brossier2004}, and provides a basis for automatically collating event databases for compositional and information retrieval applications \citep{Schwarz2003}. Throughout the present document, we adopt a signal processing approach, striving to detect magnitude changes, harmonic changes and pitch leaps. Onset detection task can be roughly divided into three different blocks : 1) sound representation ; 2) computation of a detection function, and 3) extraction of note events. We will put a special focus on the first operation of investigating the most appropriate sound representation for onset extraction. A widely used approach to onset detection in the frequency domain is the spectral flux \citep{Bello2005}, where sudden changes of the sound spectrum are detected by differentiating the signal's short-time successive spectra. Detection of onsets in a monophonic signal is not a difficult problem, especially if onsets are prominent, which is the case of most ``decay" instruments \footnote{These musical instruments are defined as an instrument with a fast transient attack, followed by an exponential release behavior, depending on the free-resonating properties of the instrument. As opposed to ``sustain" instruments, defined as the ones characterized by having a constant energy/timbre behavior over the note duration (almost like woodwinds, strings, organ, but even with less attack timbre variations).}. For example, onsets in a monophonic piano signal could be calculated with high accuracy by simply locating peaks in the amplitude envelope of the input signal. However, in complex polyphonic mixtures of music, there are simultaneously occurring events with different combinations of playing techniques. Because the amplitude envelope of an entire signal reveals little of what is going on in individual frequency regions of the signal, where note onsets and offsets may occur, resulting in masking effects and blurred note transitions, polyphonic music makes it hard to detect individual onsets. As a consequence, detection functions were proposed that analyze the signal in a band-wise fashion to extract transients occurring in certain frequency regions of the signal \citep{Scheirer1998,Masri1996b}. 

On the other side, instrument recognition is defined as the automatic process of identifying an instrument given a sound input produced by it. The goals of automatic recognition of instruments are multiple: first, to provide labels for monophonic recordings, for "sound samples" inside sample libraries, or for new patches created with a given synthesizer, and to provide indexes for locating the main instruments that are included in a musical mixture. The majority of research on the automatic recognition of musical instruments until now has been made on isolated notes or on excerpts from solo performances, depending on the application. A comprehensive review of proposed approaches on instrument recognition can be found in \citep{Herrera-Boyer2003}. Common methods combine tools from audio signal processing and machine learning. A successful combination was Mel-Frequency Cepstral Coefficients (MFCCs) as a sound representation and Support Vector Machine (SVM) for the classification, as in \citep{Marques1999} who used 16 MFCCs on 0.2 second sound segments to label 8 solo instruments playing musical scores from well-known composers. Other promising applications of SVM that are related to music classification but are not specific to music instrument labelling can be found in \citep{Li2000,Whitman2001,Guo2001}. Sophisticated Gaussian-Mixture and Hidden-Markov models \citep{Brown1999,Marques1999} have also been developed on such feature vectors to optimize audio classification. As other noticeable research works on instrument recognition, \citep{Eggink2004} developed a system to recognize solo instruments in accompanied sonatas and concertos, using the relative magnitudes of harmonics (normalized to the overall magnitude of the harmonics) as the feature for classification, and showed that the feature is robust against background accompaniment. \citep{Kitahara2006} developed the software Instogram for instrument recognition, based on the temporal trajectory of instrument existence probabilities for every possible $F_0$.

Over the past few years much research has been devoted to finding effective representations of sound to address the many challenging problems in MIR \citep{Mirex2011}. Common sound representations include Fourier transform, MFCCs and Wavelets, which have all been extensively used in MIR tasks. Extending these representations, \citep{Mallat2012} developed a mathematical operator called deep scattering transform, consisting in a cascade of wavelet decompositions and modulus operators. On the contrary to common wavelet transforms \citep{Daudet2001}, scattering transform was developed around the notion of invariability, crucial to define robust instrument-specific templates used in supervised classification. All current application studies have revealed a great potential of scattering representation to serve automatic retrieval/classification tasks of unidentified numerical objects, including musical instrument classification \citep{Anden2011},  note characterization \citep{Anden2012,Anden2014}, genre recognition \citep{Chen2013}, face recognition \citep{Chang2012}, environmental sounds \citep{Bauge2013}, texture classification \citep{Bruna2013} and medical signal analysis \citep{Chudacek2013}. Our paper aims to build upon these achievements, and to further strengthen these experimental validations through the two AMT subtasks of onset detection and instrument recognition.

We now present the organization of this paper. In section \ref{TheorSR}, we present briefly the theoretical background of scattering transforms, tracing its evolution through well-know sound representation methods. In section \ref{Methods}, we develop methods and computational details for the two tasks of onset detection and instrument recognition. In section \ref{Results}, the performances of scattering method are evaluated comparatively to other methods, and discussed to highlight the potential usefulness of scattering for some AMT applications.

\section{Sound representation : from Fourier to Scattering transforms}\label{TheorSR}

\subsection{Fourier transform}

The short-time Fourier transform of a time series $x$, used to build classical spectrogram, is defined by

\begin{equation}
\hat{x}_{t,T}(\omega) = \int x(u)w_{T}(u-t) e^{-i \omega u} du
\end{equation}

where $w_T$ is a time window of size T. Spectrograms compute then locally time-shift invariant descriptors over durations limited by a window. However, \citep{Anden2014} showed that high-frequency spectrogram coefficients are not stable \footnote{Stability means that small signal deformations produce small modifications of the representation, measured with a Euclidean norm. This is particularly important for classification.} to variability due to time-warping deformations, which often occur in audio signals. 

\subsection{Mel-Frequency spectral transform}

Mel-Frequency Spectral Coefficients (MFSCs) are obtained by averaging the spectrogram $|\hat{x}_{t,T}(\omega)|^2$ over mel-frequency intervals, which can be written as 

\begin{equation}\label{MFSCexp}
MFSC_{s}x(t,j) = \frac{1}{2 \pi} \int |\hat{x}_{t,T}(\omega)|^2 |\hat{\psi}_j (\omega)|^2 d\omega
\end{equation}

where each $\hat{\psi}_j$ covers a mel-frequency interval indexed by j. MFFCs are cosine transforms of MFSCs, and are efficient local descriptors at time scales up to T ($\approx 25 ms$). Log-frequency scales have actually been widely used in various audio processing applications, because it makes signals stable to deformation by low-pass filtering it, hence removing the more variable high-frequency content. To capture longer-range structures, these MFCCs are either aggregated in segments \citep{Bergstra2006} that cover longer time intervals, or are complemented with other features such as Delta-MFCCs \citep{Furui1986}.

\subsection{Wavelets}

The Continuous Wavelet Transformation (CWT) was introduced in order to overcome the limited time-frequency localization of the FFT for nonstationary signals and was found to be suitable in a lot of applications \citep{Kronland-Martinet1987}. Unlike the FFT, the Continuous Wavelet Transformation has a variable time-frequency resolution grid with a high frequency resolution and a low time resolution in low-frequency area and a high temporal/low frequency resolution on the other frequency side. In that respect it is similar to the human ear which exhibits similar time-frequency resolution characteristics \citep{Tzanetakis2001}. Mathematically, wavelet analysis seeks to address the defect of the Fourier  transform by decomposing the time series into local, time-dilated, and time-translated wavelet components using time-frequency atoms or wavelets $\psi$: \begin{equation}
FWT(a,b) = \frac{1}{\sqrt{a}} \int_{\infty}^{\infty} f(t) \psi(\frac{t-b}{a}) dt
\end{equation} 

$\psi(.)$ is the basic wavelet function that satisfies certain very general condition, a is the scale and b is the time shift. $FWT(a,b)$ is then the ``energy" of f(t) of scale a at time b. 

%Wavelet analysis is attractive because (1) it is local; (2) it has uniform temporal resolution for all frequency scales; (3) it is useful for characterizing gradual frequency changes. However, the optimal choice of the wavelet family, given a sound database, remains a difficult problem.

\subsection{Scattering representation}\label{ScattTheo}

Scattering representation was introduced by \citep{Mallat2012}. It is interesting to underline the motivations which originate this method, conceived as a wavelet-based extension of MFSCs. We already mentioned that high-frequencies are more sensitive to deformation than low-frequencies, which makes the Fourier-based spectrogram particularly non-adapted to take into account small deformations of a signal. The logarithmic averaging used in a mel-scale removes this instability, providing to the MFSCs a non-variable representation of a signal from one observation to another. However, this averaging also induces a loss of high-frequency information, especially over time intervals larger than T, which is why mel-frequency spectrograms are limited to such short time intervals. Based on these observations, scattering aims to provide a stable transform, yet without any loss of information.

In mathematical terms, we saw that MFSCs coefficients are defined as the spectrogram averaged along a mel-frequency scale. The averaging resulting from the integration of the MFSCs (eq. \ref{MFSCexp}) is actually equivalent (by the Parseval's theorem) to convolute them with a low-pass filter $\phi_j(t)$, as follows $|x * \psi_J(t)| * \phi_j(t)$. This averaging naturally losses high-frequency information. The scattering transform then aims to recover the information lost by averaging, observing that equation $|x * \psi_{j_1}(t)| * \phi_J(t)$ can be written as the low-frequency component of the wavelet transform of $|x * \psi_{j_1}(t)|$, i.e. : \begin{equation}
W  x * \psi_{j_1}(t) =
\left( \begin{array}{c}
 |x * \psi_{j_1}| * \phi_J(t)  \\
||x * \psi_{j_1}| * \psi_{j_2}| * \phi_J(t) \\
\end{array} \right)_{j_2 < J+P}
\end{equation}

where J and P delimit filter supports on which each wavelet $\psi$ is dilated in a specific way (see \citep{Anden2011} for details). Since the wavelet transform is invertible, the information lost by the convolution with $\phi_J$ is recovered by the wavelet coefficients $|x * \psi_{j_1}| * \psi_{j_2}$. Averaging $|x * \psi_{j_1}| * \psi_{j_2}$ by  $\phi_J$ again entails a loss of high frequencies, which can be recovered by a new wavelet transform. Iterating this process leads to establish the scattering transform S, stable to deformation, \begin{equation}\label{EqCasWav}
S x(t) =
\left( \begin{array}{c}
 x * \phi_J(t)  \\
|x * \psi_{j_1}| * \phi_J(t)  \\
||x * \psi_{j_1}| * \psi_{j_2}| * \phi_J(t)  \\
\vdots \\
||\hdots | x * \psi_{j_1}| * \psi_{j_2}| * \phi_J(t)  \\
\end{array} \right)_{j_1,j_2, ... < J+P}
\end{equation}

\begin{figure}[htbp]
\begin{center}
\resizebox{9cm}{!}{\includegraphics{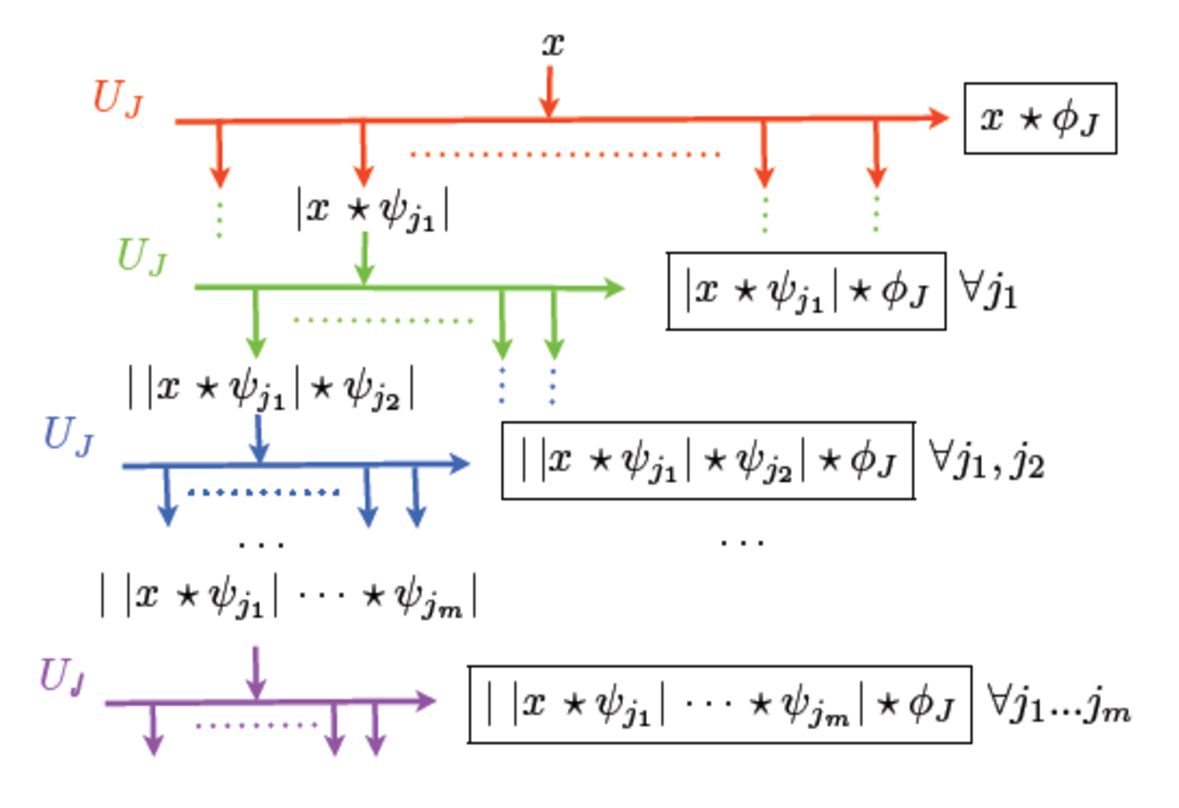}}
\caption{A scattering operator is a cascade of wavelet modulus operators $U_J$, outputting convolutions with $\phi_J$ as shown in boxes (from \citep{Anden2011}).}
\label{DiagBlocScatt}
\end{center}
\end{figure}

%\citep{Anden2011,Anden2014} showed that MFSCs could actually be interpreted as a wavelet decomposition, defined by the following filter bank
%
%\begin{equation}
%W_{t,T}x =
%\left( \begin{array}{c}
% x * \phi_J(t)  \\
% x * \psi_j(t)  \\
%\end{array} \right)_{j < J+P}
%\end{equation}
%
%where $\phi_J$ is a low-pass filter (see \citep{Anden2011} for details on these parameters). 

The calculation of eq. \ref{EqCasWav}, as illustrated in figure \ref{DiagBlocScatt}, can be viewed as a cascade of a wavelet modulus propagator U, defined as U x = $\{x * \phi,|x * \psi_{\lambda}| \}_{\lambda \in \Lambda}$. That is why, the scattering representation finds its main application in characterizing high-frequency acoustic features. While providing a multiscale representation of x, the scattering transform consists of a highly non linear transform, as opposed to the underlying discrete wavelet transform. Furthermore, one can also separate a filter factor from an excitation factor using a logarithm and a discrete cosine transform, an operation similar to MFCCs. Such coefficients are referred to as Cosine Log Scattering Coefficients (CLSCs). Still like in MFCCs, only the low-frequency of the discrete cosine transform coefficients are kept.

%Figure \ref{ScatteringIllustration} represents the three levels of CLSCs on a musical piece of piano. 

%\begin{figure}[htbp]
%\begin{center}
%\resizebox{12cm}{!}{\includegraphics{ScatteringIllustration}}
%\caption{Three levels of scattering coefficients on a musical piece of piano.}
%\label{ScatteringIllustration}
%\end{center}
%\end{figure}

The scattering feature vector for each time position k is then defined as $S x(k)=( { S x(j_1,k) }_{1 < j_1 < J} , { \tilde{S} x(j_1,j_2,k) }_{1 < j_1 < j_2 < J})$, where only the two first order coefficients are exploited, and $\tilde{S} x(j_1,j_2,k) = \frac{S x(j_1,j_2,k)}{S x(j_1,k)}$ represents the normalized second order scattering coefficients, so as to remove the dependency of the amplitude of second order coefficients upon that of the first order coefficients. The feature dimensionality here is 386.
  
In practice, in the present contribution, a complex wavelet is used, consisting of the analytic part (restriction to positive frequencies) of a Battle-Lemari'e cubic spline wavelet \citep{Mallat2012}. The window $\phi$ is the cubic spline scaling function associated to this wavelet. For all scattering computation, we use the ScatNet MATLAB software, available at \url{http://www.di.ens.fr/data/software/scatnet/}.

%Applying Parseval' theorem to eq. \ref{MFSCexp} \begin{equation}
%M_{FSCs}x(t,j) = \frac{1}{2 \pi} \int |x_{t,T} * \psi_j(u)|^2 du
%\end{equation}

\section{Methods}\label{Methods}

\subsection{Methods for onset detection}

\subsubsection{Onset detection function}\label{DefODF}

The aim of Onset Detection Function (ODF) is to highlight onsets in the signal so as to provide a clear onset trace. We use a simple energy-based ODF built upon the spectral flux. The spectral flux \citep{Masri1996} describes the temporal variation of the magnitude spectrogram by computing the difference between two consecutive short-time spectra. This ODF gives a measure of the non-stationarity of the signal in each frame of the spectral transform by calculating the deviation of each frequency bins' energy from a prediction made using the previous frames. In mathematical terms, we have

\begin{equation}
D(n) = \sum_{k=1}^{k = N/2} H(|X(n,k)| - |X(n-1,k)|)
\end{equation}

with $H(x) = \frac{x + |x|}{2}$ being the half-wave rectifier function. The rectification has the effect of counting only those frequencies where there is an increase in energy, and is intended to emphasize onsets rather than offsets \citep{Duxbury2003}. We have chosen to omit other common methods such as phase deviation \citep{Bello2004}, high frequency content \citep{Masri1996} or rectified complex domain\footnote{The main reason being that we only address decay instruments, which generally present sharp onset phases, although a certain variability can be obtained in onset shapes depending on playing techniques, e.g. excitation mode and amplitude.} \citep{Dixon2006}, since they only exhibited negligible enhancements on our test experiments.

\subsubsection{Peak picking}\label{DefPP}

The shape of the ODF bears a great importance. In an ideal case, at those time instants where phenomenal accent occur the ODF would display well-localized narrow peaks whose magnitudes are proportional to the sound intensity change. A simple peak-picking above a fixed threshold would be enough to find onset locations. In practice, the ODF tends to be much noisier over the range of real world signals for a number of reasons, such as the occurrence of various onset types, a low signal-to-noise ratio and loudness variations. To take into account these variations, most ODFs are post-processed with a dynamic thresholding $\delta(n)$, and in our case the corresponding activation probabilities, which can be computed (adapted from \citep{Essid2005}) as 

\begin{equation}\label{DefThres}
\delta(n) = \delta_{static} + median( p(n-M)), ... , p(n+M) )
\end{equation}

with $\delta_{static}$ an offset coefficient. This threshold $\delta$ is applied to the onset function, leading to a thresholded observation, whose non-zero values indicate peaks in the STFT, which can be simply picked with a maxima search. The peak-processing stage selects onset candidate peaks above the adaptive threshold $\delta$ and discards those being too small in a 25 ms range around a larger peak.

\subsubsection{Evaluation procedures}\label{AlgoOnsetDet}

To evaluate comparatively the performances of scattering representation on the task of onset detetion, the different sound representations presented in section \ref{TheorSR} (i.e. FFT, MFCC, Wavelets) are also applied to this task with the same procedure exposed in sections \ref{DefODF} and \ref{DefPP}. Also, tu put into perspective our results, two AMT algorithms are used considered : Tolonen model \citep{Tolonen2000} and the HALCA algorithm \citep{Fuentes2012}. HALCA is a state-of-the-art algorithm for the task of onset detection, evaluated by the MIREX community with an average transcription score of 62 \% \citep{Benetos2013d}, and will serve as a performance benchmark for all results.

Evaluation of onset detection is naturally performed using a note-oriented approach \citep{Fonseca2009}. We then define correct note events based on tolerance errors around onset estimations. A correct note event is then assumed to be correct if its onset is within a 40 ms range of a ground-truth onset. Evaluation metrics are defined by following equations \ref{Metr1}-\ref{Metr4} \citep{Bay2009,Mirex2011}, resulting in the note-based onset-offset recall (TPR), fall-out (FPR), precision (PPV) and F-measure (the harmonic mean of precision and recall) :

\begin{equation}\label{Metr1}
\text{TPR} = \frac{\sum_{n=1}^N {\text{TP}[n]}}{\sum_{n=1}^N {\text{TP}[n] + \text{FN}[n]}} \end{equation}
\begin{equation}\label{Metr2}
\text{FPR} = \frac{\sum_{n=1}^N {\text{FP}[n]}}{\sum_{n=1}^N {\text{FP}[n] + \text{TN}[n]}} \end{equation}
\begin{equation}\label{Metr3}
\text{PPV} = \frac{\sum_{n=1}^N {\text{TP}[n]}}{\sum_{n=1}^N {\text{TP}[n] + \text{FP}[n]}} 
\end{equation}
\begin{equation}\label{Metr4}
\text{F-measure} = \frac{2 . \text{PPV} . \text{TPR}}{\text{PPV}+\text{TPR}} 
\end{equation}

where N is the total number of notes, and TP, FP and FN scores stand for the well-known True Positive, False Positive and False Negative detections. 

The TPR and FPR metrics (eq. \ref{Metr1}-\ref{Metr2}) can then be included in ROC (Receiver Operating Characteristic) curves, which is a graphical plot illustrating the performance of a binary classifier system as its discrimination threshold is varied \citep{Kay1998}. For each value, indexed by the entire k, of the scaling factor $C_k$ present in the threshold $\delta(n)$ (eq. \ref{DefThres}), 20 different musical sequences are tested and the average of their respective metrics is used as the coordinate $d_k$ associated to this value. Along a ROC curve, we can define the so-called operation point, defined as the threshold position giving the best detection score, i.e. the highest TPR and lowest FPR. Based on the $TPR_{OP}$ and $FPR_{OP}$ scores at this point, numerical performances can be attributed for each method through the metric $E_{OP}$ defined as 

\begin{equation}\label{Topti}
E_{OP} = \sqrt{(1-TPR_{OP})^2 + FPR_{OP}^2}
\end{equation}

\subsection{Methods for instrument recognition}

\subsubsection{Support Vector Machine}

Based on the structural risk minimization inductive principle \citep{Vapnik1998}, a SVM is machine learning technique using a systematic approach to find a linear function with the lowest complexity. For linearly non-separable data, SVMs can (non-linearly) map the input to a high dimensional feature space where a linear hyperplane can be found. This mapping is done by means of a so-called kernel function. The classification performances of feature vectors are evaluated with an SVM classifier computed with a Gaussian kernel.

\subsubsection{Evaluation procedures}

For instrument recognition, we formatted our scattering coefficients by log-compressing their coefficients in order to reduce intra-instrument variation, getting the already defined CLSCs (see Sec. \ref{ScattTheo}), and by removing redundancy between coefficients with a Principal Component Analysis pre-processing, shifting down their dimension from 346 to 50, comparable to the MFCC order. CLSCs will be evaluated comparatively to the Delta-MFCCs coefficients \footnote{Basic MFCC coefficients have been proved to be outperformed by delta-MFCC in \citep{Anden2011}, which was also the case in our simulation experiments, so we skipped these sound representations.} in the task of instrument recognition. Delta-MFCCs \citep{Furui1986} are defined as the difference between MFCC coefficients of two consecutive audio frames and thus cover a time interval of twice the size. These complement the ordinary MFCCs, providing information on the temporal audio dynamics over longer time intervals. 

Each audio track of our evaluation datasets (being either note samples or musical excerpts) is decomposed in frames of duration T which are represented using Delta-MFCCs or CLSCs. Our instrument recognition has been made regardless of the pitch of sound samples. A multi-class SVM is implemented over the audio frames with a 1 vs 1 approach which trains an SVM to discriminate each pair of classes. To classify a whole track, each frame is classified using the SVM and the class with the largest number of frames in the track is selected (a method called Maximum voting). The Gaussian kernel parameter  and the SVM slack variable C are optimized with a cross-validation on a subset of the training set. For the classification part, the dataset was randomly split into training (70\%) and test data (30\%). The training data were used to train the classifier, which was then tested with a five-fold cross-validation with the unseen test data to give success rates for each class, expressed as a confusion matrix. Final results are calculated in the form of the error rates of wrongly classified tracks in the test set.

%, as well as with a more simple method based on the energy distribution in the spectral envelop \citep{Kitahara2006}, computed from harmonic magnitudes, and label HarMag in the following

\subsection{Sound datasets}

Table \ref{Table_DetailsDatasound_scatt} provides full details on the sound material used for evaluation of the tasks of onset detection and instrument recognition in this paper, and table \ref{Table_DetailsDatasound_DatasetFormat} details how the different datasets are defined. Before detailing specific datasets for the two tasks of onset detection and instrument recognition, we detail their common characteristics. For each instrument and pitch, 18 sound templates were extracted, from 3 different instrument models. Sources are detailed in table \ref{Table_DetailsDatasound_scatt}. The continuous musical excerpts are 23.8s-long\footnote{The scattering code used requires the input signal to have a power of 2 number of samples.} sequences randomly selected from different musical pieces (resulting either from MIDI scores or real audio recordings). We remove from the generated sequences the truncated notes.

\subsubsection{Onset detection}

For evaluation of the onset detection task, we only used instruments $I_1$ and $I_2$, namely the \textit{marovany} zither and the classical piano. A first dataset $D_{OD}^1$ was built automatically by combining MIDI scores and pre-recorded real instrument samples per pitch \footnote{For the \textit{marovany} repertoire, these scores are not properly speaking MIDI scores, but present a pianoroll-like representation with the same information for each note, although these information are not discretized but kept continuous. For sake of simplicity, we keep the term of MIDI for these scores.}. Such a generative process allows generating a large set of labelled training data with a minimal amount of human labour, allowing the inclusion of an important variety of different instruments and scores. In addition to that, it uniformizes the different test sequences, removing variability between data due to recording and production conditions. When synthesizing a sequence from a MIDI score and real note templates, an instrument model is first selected. Then, for each note location, the template is scaled to the correct amplitude given by the MIDI file for each note event. As our templates encompass a certain variety in the playing dynamic, we can match as well as possible the modifications of timbre induced by this dynamic. To do so, our templates are first ranged by amplitudes, and selected accordingly to their position on the MIDI amplitude scale. Our template-based MIDI sequences were also degraded using the Audio Degradation Toolbox \citep{Mauch2013}, to approximate audio recording conditions. One may criticize that they are less realistic than real recordings, but actually this process is not that far different from other MIDI-driven dataset generation such as the Disklavier technology (used e.g. in \citep{Poliner2007,Emiya2008}), as musical parameters are ``MIDI-discretized" all the same. Although for the piano, scores are actual MIDI files selected from a specialized webpage hosting MIDI files freely available under Creative Commons licenses, for the \textit{marovany} repertoire, scores were obtained with an original multi-sensor retrieval system \citep{Cazau2013c}. Although quite invasive as needing a complex experimental set-up during recording sessions, such a system allows for fast and very reliable polyphonic transcriptions. 

%The note distributions of all musical pieces for each dataset are displayed in figure \ref{BarDistribNotesTrainTest}.

Also, three other datasounds were derived from the dataset $D_{OD}^1$, each of which emphasizing a specific acoustic or musical feature $\beta$ controlled by a numerical parameter $\lambda_{\beta}$. We now detail these three other sound datasets : 

\begin{description}

\item[$D_{OD}^{2}$] with the SNR feature This acoustic feature is normalized on an unitary range, and then directly modified by the factor $\lambda_{SNR}$, which is itself controlled by the parameter SNR of the Audio Degradation Toolbox \citep{Mauch2013} ;

\item[$D_{OD}^{3}$] with the Sparsity degree feature This musical feature refers to the number of simultaneous and successive active notes in a 10-s time interval. Such a feature can then be controlled by the factor $\lambda_{POL}$ in the construction of our synthetic sequences, by simply forcing the number of notes to be lower than this factor in 10-s intervals ; 

\item[$D_{OD}^{4}$] with the Intermodulation feature This feature is characterized by strong modulations taking the form of peaks and valleys in the temporal envelop. It depends both on the intrinsic timbre properties of an instrument, and the playing mode of a musician. This feature generally makes onset detection trickier, as the strong modulations induced may result in fake transients with soft attacks. We quantified the intermodulation strength $\lambda_{INT}$ of a given note sample with the Amplitude Modulation acoustic descriptor (see \citep{Peeters2011}), and selected a template during the generation of a musical sequence accordingly to the desired $\lambda_{INT}$ value. This feature can be very prominent in repertoires of plucked string instruments, and especially in the \textit{marovany} repertoire. 
\end{description}

Eventually, to put into perspective our results on the dataset $D_{OD}^1$ with a more realistic dataset, we also used the sound dataset $D_{OD}^{5}$ which gather the same musical pieces as in $D_{OD}^1$, although now they are extracted from real world recordings.

% TABLE ONLINE !!!! Table_DetailsDatasound_scatt
\begin{table}[h]
\begin{center}
\resizebox{14cm}{!}{
\begin{tabular}{|c|c|c|c|c|}
\hline
Instrument             & Labels & Sound samples             & MIDI score                     & Audio recordings                                 \\ \hline
Marovany               & $I_1$  & Personal recording in lab & Multi-channel retrieval system & Personal recording in lab                                    \\ \hline
Piano                  & $I_2$  & MAPS (ref : 111-112-113)  & Classical Piano MIDI page      & 100 Best Piano Classics (ed. Warner Classics)                \\ \hline
Classical nylon guitar & $I_3$  & RWC (ref : 91-92-93)      & X                              & The Best Of Classical Guitar (ed. GHA)                       \\ \hline
Acoustic guitar        & $I_4$  & RWC (ref : 111-112-113)   & X                              & 20 Best of Accoustic Guitar (ed. Three Sides Now)            \\ \hline
Banjo                  & $I_5$  & RWC (ref : 361-362-363)   & X                              & Solo Banjo Works (art. Bela Fleck and Tony Trischka)         \\ \hline
Harpsichord            & $I_6$  & RWC (ref : 31-32-33)      & X                              & Art of the Baroque Harpsichord (art. Cummings L., ed. Naxos) \\ \hline
Mandolin               & $I_7$  & RWC (ref : 121-122-123)   & X                              & Czech it out (art. Zenkl, R.)                                \\ \hline
Koto                   & $I_8$  & RWC (ref : 381-382-383)   & X                              & The Art of the Koto (art. Yoshimura N. )                     \\ \hline
\end{tabular}
}
\end{center}
\caption{Details of our sound material, with the sources of the sound samples, MIDI score (only needed for the \textit{marovany} and the classical piano) and the audio recordings. The term art. stands for artist and ed. for editors.}\label{Table_DetailsDatasound_scatt}
\end{table}

\subsubsection{Instrument recognition}

Our evaluation of the instrument recognition task processing will also use two different sound datasets, so as to extend our results to different scenarios of applications. A first one labelled $D_{IR}^1$ using isolated note samples of 7 different plucked-string instruments and one piano, in order to perform note wise classification. Our instrument sound samples were all extracted from the RWC database \citep{Goto2003}, as detailed in table \ref{Table_DetailsDatasound_scatt}. During the frame-wise segmentation, windows consisting of silence signal were detected thanks to a heuristic approach based on power thresholding then discarded. Note that there is a very important trade-off in endorsing this isolated-notes strategy: we gain simplicity and tractability, but we lose contextual and time-dependent cues that can be exploited as relevant features for classifying musical sounds in complex mixtures. So in a second sound dataset labelled $D_{IR}^2$, we compiled excerpts of real recordings available commercially.

\begin{table}[h]
\begin{center}
\resizebox{13cm}{!}{
\begin{tabular}{|c|c|}
\hline
Datasets   & Description                                                                                                                                                \\ \hline
$D_1^{OD}$ & \begin{tabular}[c]{@{}c@{}}Musical sequences synthesized with MIDI scores and \\ real sound samples of marovany and classical piano\end{tabular} \\ \hline
$D_2^{OD}$ & Like $D_1^{OD}$, but parametrized with the SNR feature                                                                                                     \\ \hline
$D_3^{OD}$ & Like $D_1^{OD}$, but parametrized with the sparsity degree feature                                                                                        \\ \hline
$D_4^{OD}$ & Like $D_1^{OD}$, but parametrized with the intermodulation feature                                                                                          \\ \hline
$D_5^{OD}$ & Real musical sequences of marovany and classical piano                                                                                                     \\ \hline
$D_1^{IR}$ & Sound samples of 8 different plucked-string instruments and classical piano                                                                                \\ \hline
$D_2^{IR}$ & Real musical sequences of 8 different plucked-string instruments and classical piano                                                                       \\ \hline
\end{tabular}
}
\end{center}
\caption{Details of our sound datasets.}\label{Table_DetailsDatasound_DatasetFormat}
\end{table}

%We further split each sound sample into a transient part and a stationary part, to provide an insight into which intervals of a note might be the most useful for instrument recognition. 

\section{Results and discussion}\label{Results}

\subsection{Onset detection}

\begin{figure}[htbp]
\centering
\resizebox{12cm}{!}{\includegraphics{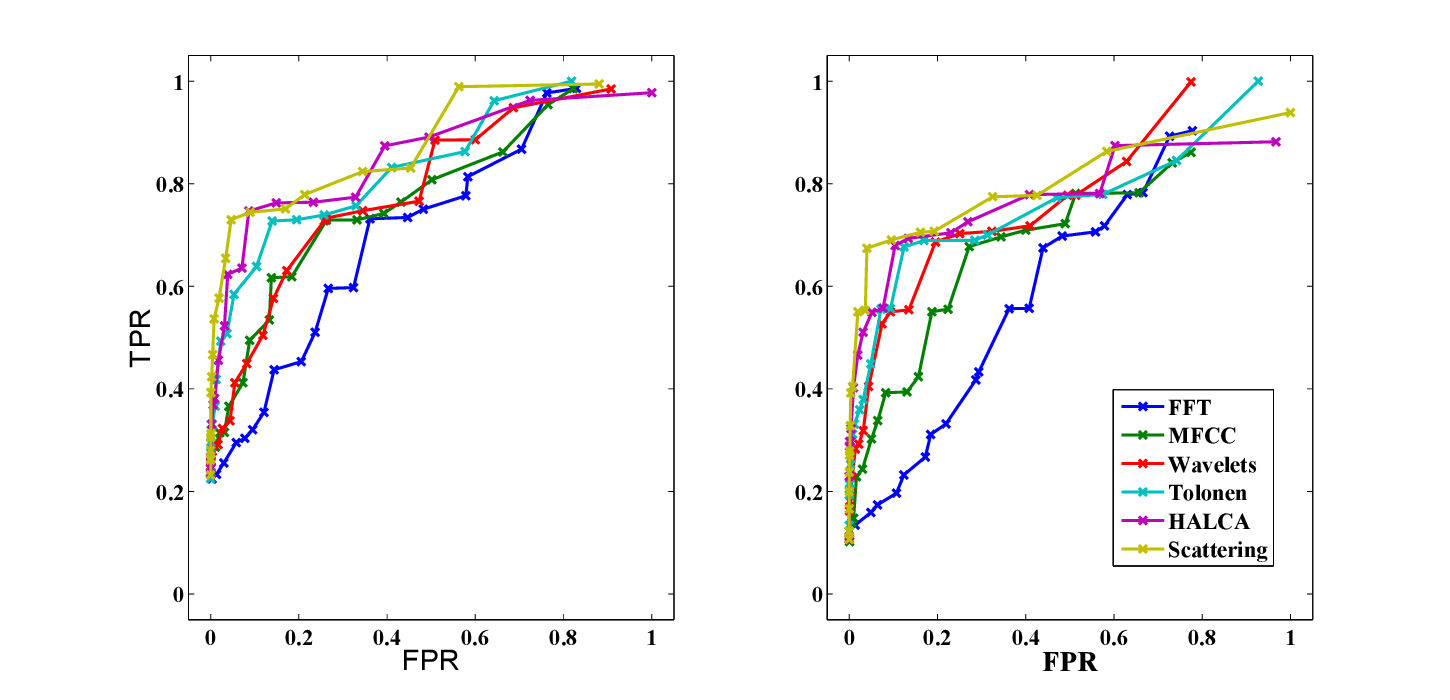}}
\caption{ROC curves for onset detection on musical sequences synthesized with real samples ($D_1^{OD}$) of the piano instrument, on the left, and for the \textit{marovany} instrument on the right.}\label{ROCcurves_OnsetDetection_SI}
\end{figure}

\begin{figure}[htbp]
\centering
\resizebox{12cm}{!}{\includegraphics{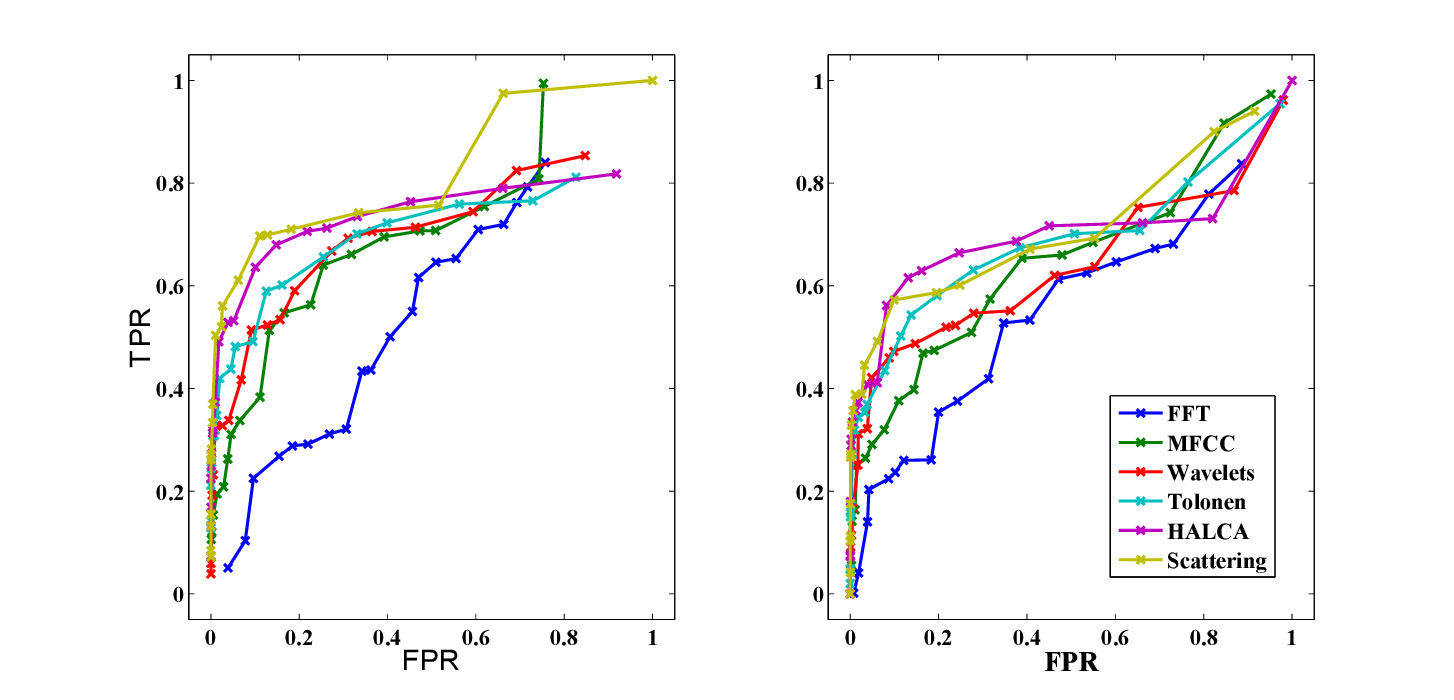}}
\caption{ROC curves for onset detection on real world musical sequences ($D_5^{OD}$) of the piano instrument, on the left, and for the \textit{marovany} instrument on the right.}\label{ROCcurves_OnsetDetection_RW}
\end{figure}

Figures \ref{ROCcurves_OnsetDetection_SI} and \ref{ROCcurves_OnsetDetection_RW} show ROC curves for onset detection of the different tested algorithms, for the sound datasets detailed in section \ref{Table_DetailsDatasound_DatasetFormat}. Considering first the synthetic database, the CLSCs achieve results close to those obtained with state-of-the-art algorithms. Numerically, for the synthesized sequences of dataset $D_1^{OD}$, we obtain the following scattering scores ($TPR_{OP}$=0.74 ; $FPR_{OP}$=0.08 ; $E_{OP}$= 0.27 for the piano and $TPR_{OP}$=0.69 ; $FPR_{OP}$=0.09 ; $E_{OP}$= 0.32 for the \textit{marovany}) and HALCA scores ($TPR_{OP}$=0.76 ; $FPR_{OP}$=0.15 ; $E_{OP}$= 0.28 for the piano and $TPR_{OP}$=0.68 ; $FPR_{OP}$=0.1 ; $E_{OP}$= 0.34 for the \textit{marovany}). The lower optimal threshold obtained for scattering operators supports the fact that this representation flattens out the values around onset peaks, and allows for a more efficient selection of peaks representative of onsets. 

For the real recording database $D_5^{OD}$,  we obtain the following scattering scores ($TPR_{OP}$=0.7 ; $FPR_{OP}$=0.11 ; $E_{OP}$= 0.32 for the piano and $TPR_{OP}$=0.57 ; $FPR_{OP}$=0.09 ; $E_{OP}$= 0.44 for the \textit{marovany}) and HALCA scores ($TPR_{OP}$=0.68 ; $FPR_{OP}$=0.15 ; $E_{OP}$= 0.35 for the piano and $TPR_{OP}$=0.63 ; $FPR_{OP}$=0.16 ; $E_{OP}$= 0.4 for the \textit{marovany}). The larger percentage of spurious notes is likely a result of a weaker signal-to-noise ratio and the playing dynamics, as the system often misses quietly played notes, which are masked by other louder notes or chords occurring shortly before or after the missed onset. In the task of onset detection, scattering transform appears to deliver a better information to facilitate the rudimentary processing of peak picking. Most contribution comes from the second order scattering coefficients, which has been shown \citep{Anden2014} to capture the high-frequency amplitude modulations that are note attacks.

\begin{figure}[htbp]
\begin{center}
\resizebox{12cm}{!}{\includegraphics{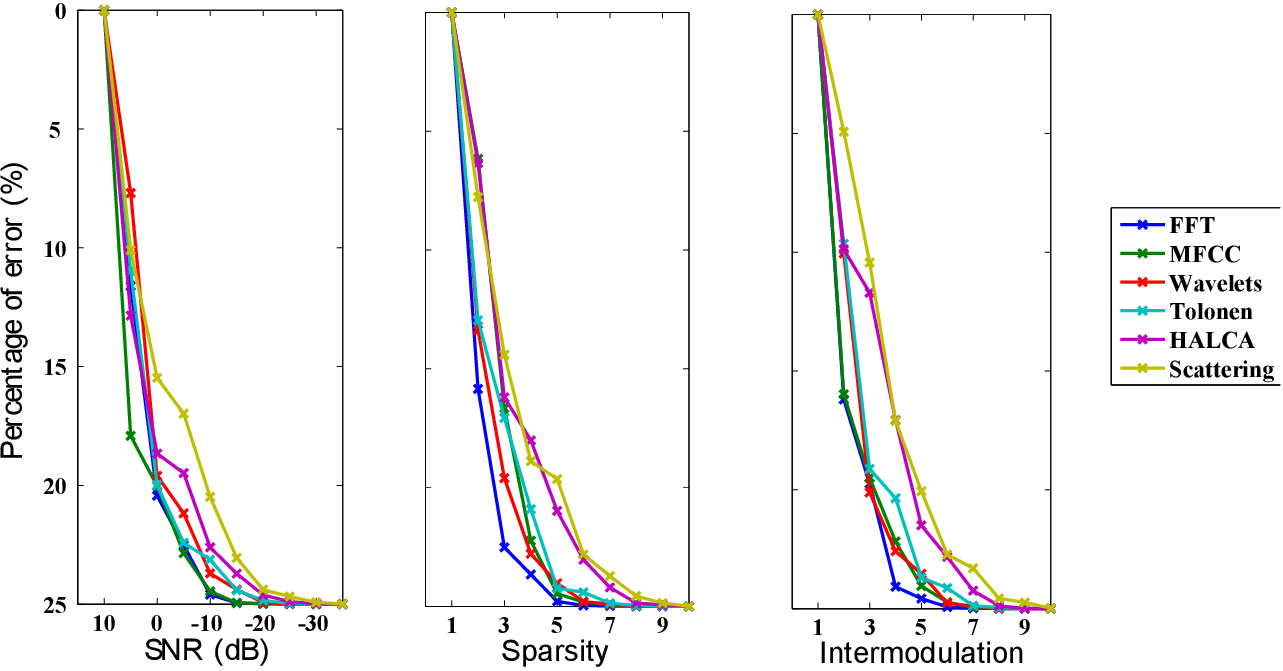}}
\caption{Relative variations (in \%) of the F-measures from the OP transcription score, for the $D_{OD}^2$, $D_{OD}^3$ and $D_{OD}^4$ datasets using the piano instrument.}
\label{ROCcurves_OnsetDetection_SensitivityAll_Piano}
\end{center}
\end{figure}

\begin{figure}[htbp]
\begin{center}
\resizebox{12cm}{!}{\includegraphics{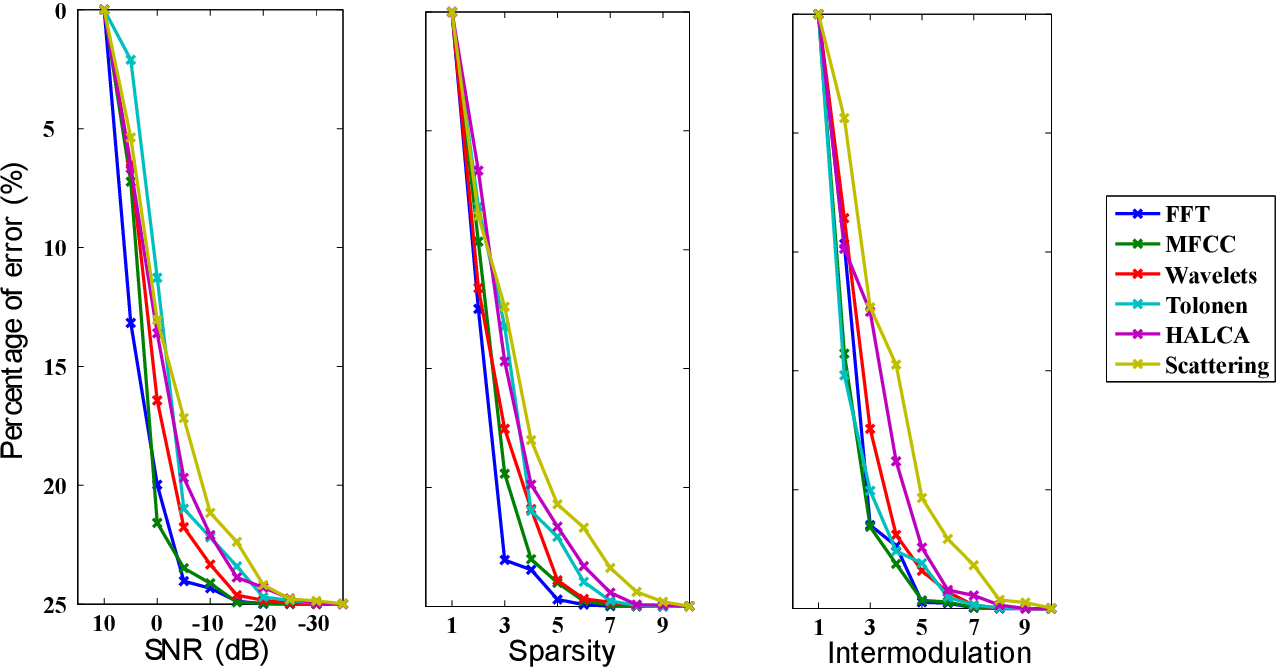}}
\caption{Relative variations (in \%) of the F-measures from the OP transcription score, for the $D_{OD}^2$, $D_{OD}^3$ and $D_{OD}^4$ datasets using the \textit{marovany} instrument.}
\label{ROCcurves_OnsetDetection_SensitivityAll_Maro}
\end{center}
\end{figure}

A more in-depth analysis is now necessary to really identify error sources. Then, as a second step in our evaluation, we tested each onset detection algorithm, calibrated on their operating point threshold, to different acoustic perturbations, namely uncorrelated noise (typically due to ambient noise and recording quality), correlated noise (which can be identified to a sparsity degree in music) and intermodulation. Figures \ref{ROCcurves_OnsetDetection_SensitivityAll_Piano}-\ref{ROCcurves_OnsetDetection_SensitivityAll_Maro} show evolution of these performances against our three acoustical features.

For all of these acoustic features of influence, scattering systematically ranks among the top performances. First, we can see that the scattering-based detection is very robust to noise, as musical features do not interfere with additive noise, which is flattened out across the different interference bands. Concerning note sparsity in music, scattering achieves also good results, showing that it can better discriminate the energy respective to each note in a given analysis frame, and also detect a transient embedded in the decay phase of a note previously played. For what concerns envelop modulation, which exhibits "valleys" in the signal waveform that can be confounded with an attack transient, scattering appears to provide superior results than other methods. An explicative comment would be that scattering discriminates more efficiently the information on envelop filtering and transient interferences than other methods. Intermodulation-related transients are smoothed out in the CLSCs representation, as a very strong high-frequency energy content must be present to reach the highest scattering level. It appears that onset detection of intermodulated notes can be done with a much more tolerant threshold descriptor than a parametric one evaluating the variations of the energy envelop. 

As a more general comment, the particular strength of scattering towards these features may be explained by the more complex and rich band-decompositions operated on high frequencies. Indeed, \citep{Scheirer1998} stated that, for onset detection, it is advantageous that the detection system divides the frequency range into fewer sub-bands as done by the human auditory system. Band-wise filtering has then been applied by many authors for this task \citep{Klapuri1999,Collins2005}. 

%The \textit{marovany} zither of Madagascar presents a certain emphasis of this feature, depending on pitches and playing modes, explaining the lower absolute performance level. 

\subsection{Instrument recognition}

Table \ref{ResultRecoInstru} gives a global overview of our results on instrument recognition. Tables \ref{ConfMat_IsoNotes} and \ref{ConfMat_ContiSeq} detail the confusion matrices of the scattering transform respective to the isolated complete note and the continuous sequence datasets. Based on our results on instrument recognition, CLSCs were shown to achieve significantly higher accuracy than other classical representations (e.g. MFCCs or Delta-MFCCs). As explicative comments about the general better results obtained with the scattering transform, we put forward its ability to recover lost non-stationary signal structures and characterize them in a richer representation (i.e. providing complementary co-occurrence information which refines MFCC descriptors), and open the possibility to capture more sophisticated auditory phenomena such as transients, amplitude and frequency modulations, time-varying filters and chord structure \citep{Anden2014}. In particular, numerous psychoacoustic studies have shown that the onset provides an important cue for timbre perception and thus musical instrument identification, particularly in the case of isolated tones \citep{Clark1964,Risset1969,McAdams1993}. As we saw that note onsets are well captured in the second-order scattering coefficients, this may explain its higher performances then MFCCs.

\begin{table}

 \begin{center}
\resizebox{14cm}{!}{
\begin{tabular}{|c|c|c|c|c|c|c|c|c|c|c|}\hline
Datasets &  Methods & $I_1$ & $I_2$ & $I_3$ & $I_4$ & $I_5$& $I_6$& $I_7$& $I_8$& \begin{tabular}[x]{@{}c@{}}Average\\(and standard deviation)\end{tabular}  \\ \hline

\multirow{2}{*}{$D_1^{IR}$} 

%& HarMag & 24.4 & 22.1 & 20.1 & 18.2 & 15.9 & 17.4 & 31.1 & 22.8 & 21.5 ($\pm$ 4.8) \\ \cline{2-10}

& Delta-MFCC & 23.3 & 21.4 & 19.9 & 18.7 & 16.3 & 19.9 & 29.2 & 21.6& 21.3 ($\pm$ 3.8) \\ \cline{2-10}

& Scattering & 19.1 & 20.8 & 23.6 & 14.1 & 16.1 & 15.6 & 26.3 & 17.7& 19.2 ($\pm$ 4.2)  \\ \hline

\multirow{2}{*}{$D_2^{IR}$} 

%& HarMag & 21.1 & 23.4 & 20.7 & 19.6 & 14.5 & 18.1& 32.3 & 22.8 &  21.3 ($\pm$ 5.5) \\ \cline{2-10}

& Delta-MFCC & 23.1 & 19.4 & 19.8 & 18.9 & 16.7 & 15.1 & 32.6 & 23.4 &  21.1 ($\pm$ 5.4)  \\ \cline{2-10}

& Scattering & 16.7 & 17.8 & 22.1 & 14.4 & 15.3 & 12.1 & 23.7 & 17.4& 17.4 ($\pm$ 3.3)  \\ \hline

\end{tabular}
}
\end{center}

\caption{Instrument recognition results given in terms of error rate classification for the datasets $D_1^{IR}$ and $D_2^{IR}$ using Delta-MFCC and scattering representations.}
\label{ResultRecoInstru}

\end{table}

When considering continuous excerpts and longer analysis segments, the error decreases as larger-scale musical information is encoded. Indeed, continuous excerpts of music includes complementary musical features of the timbre, such as rhythm and playing techniques, from which scattering transform seem to benefit the most for the task of instrument recognition. Such an observation has already been made in past studies \citep{Marques1999}. Here is pointed out a major difficulty of audio representations for classification, i.e its multiplicity of information at different time scales: pitch and timbre at the scale of milliseconds, the rhythm of speech and music at the scale of seconds, and the music progression over minutes and hours. Facing this difficulty, deep scattering has been precisely developed as a stable and invariant signal representation over time scales larger than 25 ms \citep{Anden2014}.

%\multirow{3}{*}{IsoNotes / Onsets} & HarMag & 29.3 & 27.1 & 24.9 & 29.6 & 23.4 & 28.1 & 36.7 & 29.9 & 21.5 ($\pm$ 4.8) \\ \cline{2-10}
%
%& Delta-MFCC & 24.1 & 21.3&19.7&19.4&16.3&20.4&29.9&21.8 & 21.6 ($\pm$ 4) \\ \cline{2-10}
%
%& Scattering & 19&20.5&23.7&13.4&16&15.9&27.4&17.5 & 19.2 ($\pm$ 4.6)  \\ \hline
%
%
%
%\multirow{3}{*}{IsoNotes / Stationary} & HarMag & 24.2 & 22.5 & 19.8 & 18.1 & 16 & 17.6 & 31.2 & 22.4 & 21.5 ($\pm$ 4.8) \\ \cline{2-10}
%
%& Delta-MFCC & 23.7&21.2&19.4&19.3&16.2&19.8&29.7&21.1& 21.3 ($\pm$ 4) \\ \cline{2-10}
%
%& Scattering & 19&20.3&23.4&13.8&15.8&15.8&26.2&17.8& 19 ($\pm$ 4.1)  \\ \hline

\begin{table}[!htb]

\centering

\resizebox{8cm}{!}{
\begin{tabular}{|c|c|c|c|c|c|c|c|c|}

\hline 

& $I_1$ & $I_2$ & $I_3$ & $I_4$ & $I_5$& $I_6$& $I_7$& $I_8$\\[0.5ex] \hline 
 $I_1$ & \textbf{80.9} & 3& 0.8& 1.6 & 2.7 &3.4 & 3.6 & 4 \\ 
 $I_2$ &  4.8 & \textbf{79.2}& 5.1 & 2.3& 1.6 & 1.4 & 2.5 & 3.1 \\ 
 $I_3$ &  0.5 & 1.6 &  \textbf{76.4}& 4.8 & 3.7 & 5.6 & 4.4 & 3 \\ 
 $I_4$ &  1.6 & 0.8&  3.3 & \textbf{85.9} & 1.8& 1.9 &0.8& 3.9\\ 
 $I_5$ &  1.3 & 2.2&  1.2& 3.1 & \textbf{83.9} & 2.6 &2.8& 2.9 \\ 
 $I_6$ &  1.4 & 0.5&  1.7 & 3.5 & 3.4&\textbf{84.4} & 3.8 & 1.3\\
 $I_7$ &  6.3 & 4.5&  1.4 & 1.9 & 6.9 & 1.5 &\textbf{73.7} &3.8 \\
 $I_8$ &  4.9 & 3.6& 2.4 & 2.5 & 0.3 &3.4 & 0.4 & \textbf{82.8} \\[1ex]
\hline
\end{tabular} 
        }
        \caption{Confusion matrices for scattering coefficients for the datasets of isolated complete notes.}\label{ConfMat_IsoNotes}
        
\end{table}

\begin{table}[!htb]

\centering

\resizebox{8cm}{!}{
\begin{tabular}{|c|c|c|c|c|c|c|c|c|c|}

\hline 

& $I_1$ & $I_2$ & $I_3$ & $I_4$ & $I_5$& $I_6$& $I_7$& $I_8$ \\[0.5ex] \hline 
 $I_1$ & \textbf{83.3} & 2.3& 2.6& 2 & 2.4&3.1&2.4&1.9 \\ 
 $I_2$ &  0.2 & \textbf{82.2}& 3.8 & 2.9& 4.3&2&0.7&3.9 \\ 
 $I_3$ &  3.7 & 0.6&  \textbf{77.9}& 4.7 & 2.9&3.8&3.8&2.6 \\ 
 $I_4$ &  1.5 & 0.8&  2.9& \textbf{85.6} & 0.8&2.4&3.9& 2.1 \\ 
 $I_5$ &  1.8 & 0.1&  2.3& 4.4 & \textbf{84.7} &1.9&1.8&3\\ 
 $I_6$ &  1.4 & 0.1&  2.1& 3.4 & 0&\textbf{87.9} &4.2&0.9\\
 $I_7$ &  1 & 3.5&  5.8& 5.4 & 2&4.6&\textbf{76.3} &1.4\\
 $I_8$ &  3.5 & 0.2& 2.7 & 3.2&1&3.8&3 & \textbf{82.6} \\[1ex]
\hline
\end{tabular} 
        }
        \caption{Confusion matrices for scattering coefficients for the datasets of continuous musical pieces.}\label{ConfMat_ContiSeq}
        
\end{table}

%Different results are obtained when differentiating transient and stationary parts of notes, as already observed in the previous study by \citep{Essid2005}. When considering the complete note, the information carried by the transients gets diluted over the entire signal, hence its impact on the final classification decision becomes weak.

\section{Conclusion}\label{Conc}

In this paper, we tested the use of the deep scattering transform on the two tasks of onset detection and instrument recognition, which are of first important for automatic music transcription \citep{Klapuri2004b}. Taken altogether our results based on various simulation experiments, scattering operators seem to possess certain advantages in accomplishing these tasks, relatively to other classical methods. Our two tasks of investigation benefited positively from the richer information (due to recovered high-frequency information) and the invariance property (due to successive averaging of high-frequency information) in the scattering transform. This tendency confirms the potential of scattering representations, announced theoretically by \citep{Mallat2012,Anden2012}, and already well validated by other studies in different classification/retrieval tasks (see Introduction). Eventually, larger database could be used to confirm these promising performances. Also, one noticeable drawback of the method is its high time-consuming algorithm, which would need more efficient implementation for further applications.

%The Constant Q-transform, which is a spectrogram scaled on a logarithmic axis, is commonly used on spectrogram-factorization methods such as PLCA \citep{Fuentes2011,Benetos2013}. With its richer acoustic content, it would be interesting to investigate how scattering representation could be integrated to such models.

\medskip

 \noindent \textbf{Acknowledgements}
 
 \setlength{\parindent}{0.7cm} 
 
At the risk of omitting some relevant names, the authors would like to especially thank March Chemillier (CAMS-EHESS) for recordings of the \textit{marovany} and Laurent Quartier (LAM-UPMC) for technical supports.

\end{document}